\documentclass[letterpaper]{article} % DO NOT CHANGE THIS
\usepackage[preprint]{aaai2027}
\usepackage[hyphens]{url}  % DO NOT CHANGE THIS
\usepackage{graphicx} % DO NOT CHANGE THIS
\usepackage{natbib}  % DO NOT CHANGE THIS AND DO NOT ADD ANY OPTIONS TO IT
\usepackage{caption} % DO NOT CHANGE THIS AND DO NOT ADD ANY OPTIONS TO IT
\usepackage{algorithm}
\usepackage{algorithmic}

\usepackage{newfloat}
\usepackage{listings}
\DeclareCaptionStyle{ruled}{labelfont=normalfont,labelsep=colon,strut=off} % DO NOT CHANGE THIS
\floatstyle{ruled}
\newfloat{listing}{tb}{lst}{}
\floatname{listing}{Listing}

\usepackage{booktabs}
\usepackage{amsmath, multirow, amssymb}
\title{Clinically Structured Surrogate Rewards for Post-SFT Medical Image Captioning}
\author{
    Hyun Jun Kim\textsuperscript{\rm 1}\equalcontrib,
    Heeseung Shin\textsuperscript{\rm 2}\equalcontrib,
    Changwon Lim\textsuperscript{\rm 3}\corresponding
}

\affiliations{
    \textsuperscript{\rm 1}Department of Smart Cities, Chung-Ang University,
    84 Heukseok-ro, Dongjak-gu, Seoul 06974, Republic of Korea\\
    \textsuperscript{\rm 2}Department of AI in Medicine, Chung-Ang University,
    84 Heukseok-ro, Dongjak-gu, Seoul 06974, Republic of Korea\\
    \textsuperscript{\rm 3}Department of Statistics and Data Science, Chung-Ang University,
    84 Heukseok-ro, Dongjak-gu, Seoul 06974, Republic of Korea\\
    hyunjun0615@cau.ac.kr (H. J. Kim);
    hs970416@cau.ac.kr (H. Shin);\\
    clim@cau.ac.kr (C. Lim)
}

\begin{document}

\maketitle

\begin{abstract}
Medical image captioning requires translating heterogeneous visual evidence into concise clinical descriptions, where errors in findings, assertion states, or anatomical relations can alter clinical meaning despite surface-level fluency. Sequence-level policy optimization can directly optimize complete captions, but common rewards rely on global text similarity, direct image–caption compatibility, or unordered concept overlap, leaving visual neighborhoods and clinical-claim structure implicit. We propose a clinically structured surrogate reward framework for post-SFT medical image captioning. The framework combines biomedical semantic and short-range lexical fidelity with two structured rewards: distributional image-neighborhood alignment, which matches the medical-image-bank distributions induced by reference and generated captions, and clinical graph consistency, which applies maximum-weight one-to-one matching to entities, assertion states, and typed relations. The four rewards are independently normalized within each rollout group, combined with fixed relative weights, and optimized with GDPO. Across organizer-evaluated hidden test sets for the Standard and Synthetical ImageCLEFmedical Caption tracks and three vision–language backbones, the method improves Overall, Relevance, and Factuality over matched SFT baselines in all six backbone–track combinations, with average relative gains of 3.4\%, 2.1\%, and 5.8\%, respectively. Ablations and paired diagnostics indicate that the structured rewards provide complementary signals, reducing image-neighborhood divergence and improving entity–assertion–relation consistency.
\end{abstract}

\section{Introduction}

Medical image captioning aims to translate heterogeneous visual evidence into concise and clinically meaningful descriptions. This task is difficult because medical images vary widely in modality, anatomy, acquisition protocol, and diagnostic specificity, while small wording changes can substantially alter clinical meaning. A caption may be fluent and semantically similar to a reference yet still contain an incorrect finding, assertion state, anatomical localization, laterality, or finding--anatomy relation. Consequently, global lexical or embedding-based similarity does not by itself guarantee preservation of the underlying clinical claims~\cite{jain2021radgraph,delbrouck2022semantic,delbrouck2024radgraph}.

Most vision--language models for medical captioning are adapted through supervised fine-tuning (SFT) with teacher-forced maximum-likelihood training. Although effective, this objective optimizes next-token likelihood rather than criteria defined over complete generations, creating a mismatch between token-level training and sequence-level evaluation~\cite{bengio2015scheduled,ranzato2016sequence}. Sequence-level post-training addresses this mismatch by assigning rewards to full outputs~\cite{ranzato2016sequence,rennie2017selfcritical}, but its effectiveness depends on what those rewards represent. Common rewards measure semantic similarity, lexical overlap, direct image--caption compatibility, factual consistency, or clinical concept coverage~\cite{lin2004rouge,zhang2020bertscore,codella2024medimageinsight,delbrouck2022semantic}. These signals are useful, but they generally summarize a caption through global scores. Text-based rewards may underrepresent visual grounding, direct image compatibility focuses on a single target image, and unordered concept overlap does not explicitly preserve assertion states or typed relations among clinical entities.

We therefore propose a clinically structured surrogate reward framework for post-SFT medical image captioning. The framework retains biomedical semantic and short-range lexical fidelity while introducing two complementary structured signals. Distributional image-neighborhood alignment compares the probability distributions over a fixed medical-image bank induced by the reference and generated captions, encouraging the generated caption to recover the reference-associated visual-semantic neighborhood rather than optimizing a separate target image--caption score. Clinical graph consistency represents each caption through entities, assertion states, and typed relations, and uses maximum-weight one-to-one matching to allow partial credit for semantically similar mention spans while explicitly constraining clinical types and relations. Because the four rewards differ in scale and variability, they are independently normalized within each rollout group, combined using fixed relative weights, and optimized with Group Reward-Decoupled Normalization Policy Optimization (GDPO)~\cite{liu2026gdpo}. GDPO is used as a fixed post-SFT optimizer; our contribution lies in the structured information encoded by the reward objective rather than in a new policy-optimization algorithm.

We evaluate the framework on the Standard and Synthetical ImageCLEFmedical Caption 2026 tracks~\cite{damm2026imageclef} using three vision--language backbones. For each backbone and track, the selected SFT checkpoint initializes GDPO and serves as the matched baseline under identical prompts, model-selection criteria, and hidden-test decoding. Across all six backbone--track settings, the proposed post-SFT procedure improves organizer-evaluated Overall, Relevance, and Factuality scores over the corresponding SFT models. Reward ablations and paired post-hoc diagnostics further indicate that the structured terms act as complementary signals: the optimized models reduce reference-conditioned image-neighborhood divergence and improve entity--assertion--relation consistency under the frozen reward-side representations. Accordingly, this work makes three contributions: a distributional image-neighborhood reward over a medical-image bank; a semantic one-to-one clinical graph reward for typed entities, assertions, and relations; and consistent multi-backbone, multi-track evidence supported by component ablations and structured diagnostics. These analyses characterize the intended effects of the proposed rewards rather than independently verifying complete image-grounded clinical correctness.

Our contributions are threefold:
\begin{itemize}
    \item We introduce a distributional image-neighborhood reward that aligns the medical-image-bank distributions induced by reference and generated captions, rather than directly optimizing generated captions against a single target image.
    \item We introduce a clinical graph consistency reward based on maximum-weight one-to-one matching of clinical entities, assertion states, and typed relations, combining semantic mention correspondence with explicit structural constraints.    
    \item We demonstrate consistent improvements over matched SFT baselines across three vision–language backbones and two medical-captioning tracks, and characterize the complementary roles of the reward components through ablations and paired structured diagnostics.
\end{itemize}

\section{Related Work}

\subsection{Medical Image Captioning Across Dataset Regimes}

ROCOv2 pairs medical images from biomedical literature with figure captions, UMLS concepts, and modality annotations, covering diverse modalities and anatomical regions~\cite{ruckert2024rocov2}. ImageCLEFmedical Caption builds caption-generation benchmarks on this ROCO-family setting, including Standard and Synthetical tracks~\cite{damm2026imageclef}. Unlike MIMIC-CXR and IU X-Ray, which associate chest radiographs with study-level reports, ROCO-derived benchmarks emphasize heterogeneous single-image captioning~\cite{johnson2019mimic,demnerfushman2016iu}. The settings therefore differ in data provenance, input granularity, modality coverage, and target-text function. Our work focuses on sequence-level reward design for heterogeneous single-image captioning, where rewards must accommodate diverse visual domains while preserving clinically meaningful findings and relations.

\subsection{Sequence-Level Optimization for Medical Generation}

Sequence-level optimization assigns rewards to complete outputs rather than optimizing only token-level likelihood. Self-Critical Sequence Training directly optimizes sampled captions using task-level rewards~\cite{rennie2017selfcritical}. Recent radiology report-generation methods introduce clinically oriented post-training objectives: MPO models heterogeneous radiologist preferences through multi-objective rewards~\cite{xiao2025mpo}, HiMed-RL combines token-, concept-, and semantic-level supervision~\cite{wang2026himed}, and OraPO uses atomic-fact entailment with oracle supervision for low-information rollout groups~\cite{chen2026orapo}. These methods primarily target chest-X-ray report generation. We instead study heterogeneous single-image captioning and structure supervision through distributions over a medical-image bank and one-to-one matching of typed clinical claims. GDPO independently normalizes reward components within each rollout group before aggregation~\cite{liu2026gdpo}. We use GDPO as a fixed optimizer and focus on the reward objective rather than proposing a new policy-optimization algorithm.

\subsection{Structured Caption Rewards}

Conventional captioning rewards measure lexical or semantic agreement, direct image--caption compatibility, factual consistency, or concept overlap~\cite{lin2004rouge,zhang2020bertscore,sellam2020bleurt,zha2023alignscore,kraljevic2019medcat,codella2024medimageinsight}. Direct image similarity has also been used as an RL reward for visually distinctive captions~\cite{cho2023finegrainedimagecaptioningclip}. More recent methods introduce structured supervision: SC-Captioner rewards object--attribute--relation self-correction~\cite{zhang2025sc}, ClaimDiff-RL compares typed image-verified atomic claims~\cite{li2026claimdiff}, and CCCaption separates completeness from correctness through visual coverage and hallucination verification~\cite{tang2026cccaption}. These methods mainly target natural-image captioning and rely on generic scene structures, atomic claims, or multimodal judges.

Clinical graphs provide domain-specific claim structure. RadGraph extracts clinical entities, assertion states, and relations, while RadGraph-XL expands coverage across anatomy--modality pairs~\cite{jain2021radgraph,delbrouck2024radgraph}. Graph-overlap rewards have improved factuality in chest-X-ray report generation~\cite{delbrouck2022semantic}. Our graph reward instead uses semantic maximum-weight one-to-one matching, requiring agreement in entity type and assertion state and extending matching to typed relations. Our image-neighborhood reward addresses a separate limitation of direct image compatibility by aligning reference- and candidate-induced distributions over a fixed medical-image bank. A concurrent workshop report uses direct reference, target-image, and concept-alignment rewards~\cite{kim2026gdpo}; this work replaces the latter two with distributional image-neighborhood alignment and typed clinical-graph matching, and adds multi-backbone evaluation, ablations, and paired diagnostics.

\section{Method}

\begin{figure*}[t]
    \centering
    \includegraphics[width=\textwidth]{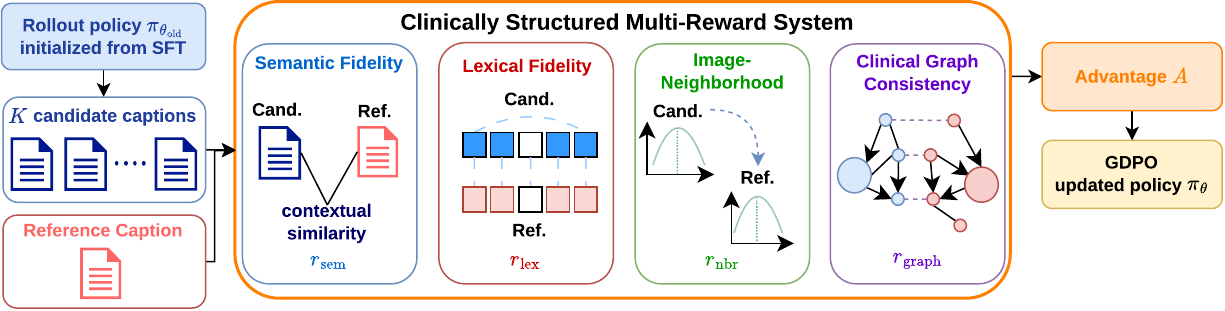}
    \caption{
    Overview of the proposed SFT--GDPO framework. The selected SFT checkpoint initializes the rollout policy and is retained as the frozen KL reference. For each input, \(K\) sampled captions are scored by four reward branches, independently standardized within the rollout group, combined using fixed weights, and converted into a caption-level advantage for GDPO.
    }
    \label{fig:architecture}
\end{figure*}

\subsection{Overview}
 Let \(\mathcal{D}=\{(x_n,y_n^*)\}_{n=1}^{N}\) denote the training set, where \(x_n\) is an image--instruction input and \(y_n^*\) is its reference caption. We use surrogate rewards as training-time proxies for clinically relevant caption properties, rather than as independent verification of image-grounded clinical correctness.

We first adapt a pretrained vision--language model through teacher-forced SFT. The selected policy \(\pi_{\mathrm{SFT}}\) initializes post-SFT optimization and is retained as the frozen reference policy for KL regularization. At each GDPO step, the rollout policy
\(\pi_{\theta_{\mathrm{old}}}\) samples \(K\) candidate captions
\[
y_{n,i}\sim\pi_{\theta_{\mathrm{old}}}(\cdot\mid x_n), \qquad i=1,\ldots,K,
\]
for each input. Each candidate receives the reward vector
\[
\mathbf{r}_{n,i}=[r_{\mathrm{sem}},r_{\mathrm{lex}}, r_{\mathrm{nbr}},r_{\mathrm{graph}}],
\]
corresponding to biomedical semantic fidelity, short-range lexical fidelity, distributional image-neighborhood alignment, and clinical graph consistency. Reference captions and reward-side models and resources are used only for training-time scoring; inference requires only the updated policy. Figure~\ref{fig:architecture} summarizes this training pipeline.

\subsection{Reference-Based Fidelity Rewards}

For readability, we write \(y\) for a generated caption and \(y^*\) for its reference.

\paragraph{Biomedical semantic fidelity.}
We use BioBERTScore-F1~\cite{lee2020biobert,zhang2020bertscore}. Let \(P_{\mathrm{sem}}\) and \(R_{\mathrm{sem}}\) denote the average maximum token-level cosine similarities from generated to reference tokens and in the reverse direction, respectively,  using normalized BioBERT representations. The reward is
\[
r_{\mathrm{sem}}=\frac{2P_{\mathrm{sem}}R_{\mathrm{sem}}}{P_{\mathrm{sem}}+R_{\mathrm{sem}}+\epsilon}.
\] 
The F1 formulation penalizes both omitted reference content and generated content without a corresponding reference match.

\paragraph{short-range lexical fidelity.}
We use a ROUGE-SU4-F1 reward~\cite{lin2004rouge}.
After lowercasing and punctuation normalization, let \(U_4(y)\)
contain all unigrams and skip-bigrams with at most four intervening
tokens. We define
\(r_{\mathrm{lex}}=F_1(U_4(y),U_4(y^*))\),
which preserves local word order and short biomedical expressions.

\subsection{Distributional Image-Neighborhood Alignment}

Let \(V=[v_1,\ldots,v_M]\in\mathbb{R}^{d\times M}\) contain the normalized MedImageInsight embeddings of all M images in the corresponding training split~\cite{codella2024medimageinsight}. Using the MedImageInsight text encoder, we map the reference and generated captions to normalized embeddings \(z^*\) and \(z\). These embeddings induce distributions over the common image bank:
\[
\begin{aligned}
p^* &= \operatorname{softmax}(V^\top z^*/\tau),\\
q   &= \operatorname{softmax}(V^\top z/\tau),
\end{aligned}
\]
where \(\tau\) is the temperature. We define the image-neighborhood reward as the negative cross-entropy
\[
r_{\mathrm{nbr}}=\sum_{j=1}^{M}p_j^*\log q_j.
\]
Because \(p^*\) is fixed within a rollout group, maximizing \(r_{\mathrm{nbr}}\) is equivalent to minimizing \(D_{\mathrm{KL}}(p^*\|q)\). The reward therefore aligns the image neighborhoods induced by the reference and generated captions instead of assigning a separate direct target image–caption score.

The full training-image bank is used. Image embeddings and reference-induced distributions are precomputed, while candidate distributions are constructed online. The bank contains no validation or hidden-test images; the current training image may appear only as an unprivileged support point.

\subsection{Clinical Graph Consistency}

We use frozen RadGraph-XL to extract clinical graphs \(\mathcal{G}(y)=(\mathcal{E}_y,\mathcal{R}_y), \mathcal{G}(y^*)=(\mathcal{E}_{y^*},\mathcal{R}_{y^*})\) from the generated and reference captions~\cite{jain2021radgraph,delbrouck2024radgraph}. These reward-side pseudo-graphs may inherit the extractor’s domain coverage and errors. An entity is represented as  \(e=(m,t,a)\), where \(m\), \(t\), and \(a\) denote its mention span, entity type, and assertion state. A relation is represented as \(\rho=(e_s,\ell,e_o)\), with source entity \(e_s\), relation type \(\ell\), and target entity  \(e_o\).

Let \(\phi(m)\) be the normalized SapBERT embedding of a normalized mention span~\cite{liu2021sapbert}. Compatibility between reference entity \(e_i^*\) and a generated entity
\(e_j\) is
\[
\begin{aligned}
s_E(e_i^*,e_j)
&= \mathbf{1}[t_i^*=t_j]\,\mathbf{1}[a_i^*=a_j] \\
&\quad \times
\bigl[\cos\!\bigl(\phi(m_i^*),\phi(m_j)\bigr)\bigr]_+,
\end{aligned}
\]
where \([u]_+=\max(0,u)\). Mention spans may therefore receive partial semantic credit, whereas entity types and assertion states must match exactly.

Let \(\mathcal{M}_E\) denote the set of valid one-to-one entity matchings. We compute
\[
W_E=\max_{\mathcal{M}\in\mathcal{M}_E}\sum_{(i,j)\in\mathcal{M}}s_E(e^*_i,e_j),
\]
\[
F_E=\frac{2W_E}{|\mathcal{E}_{y^*}|+|\mathcal{E}_y|+\epsilon},
\]
where \(|\cdot|\) denotes set cardinality. The one-to-one assignment prevents multiple generated mentions from receiving credit for the same reference entity.

For a reference relation \(\rho_u^*=(e_{u,s}^*,\ell_u^*,e_{u,o}^*)\) and a generated relation \(\rho_v=(e_{v,s},\ell_v,e_{v,o})\), we define
\[
s_R(\rho_u^*,\rho_v)=\mathbf{1}[\ell_u^*=\ell_v]\, s_E(e_{u,s}^*,e_{v,s})\, s_E(e_{u,o}^*,e_{v,o}).
\]
Using the corresponding set of valid one-to-one relation matchings \(\mathcal{M}_R\), we define
\[
W_R=\max_\mathcal{M}\in \mathcal{M}_R\sum_{(u,v)\in\mathcal{M}}s_R(\rho_u^*,\rho_v),
\]
\[
F_R=\frac{2W_R}{|\mathcal{R}_{y^*}|+|\mathcal{R}_y|+\epsilon}.
\]
The final reward is \(r_{\mathrm{graph}}=0\) if either entity set is empty, \(F_E\) if both relation sets are empty, and \(0.7F_E+0.3F_R\) otherwise. When only one caption contains relations, its unmatched relation structure receives no relation credit. Thus, \(r_{\mathrm{graph}}\) measures reference-conditioned entity–assertion–relation consistency, not complete image-grounded factuality.

\subsection{Reward Aggregation and GDPO Optimization}

Following GDPO, each reward is independently standardized within the \(K\)-caption rollout group. For group \(g\), candidate \(i\), and reward \(k\in\{\mathrm{sem},\mathrm{lex},\mathrm{nbr},\mathrm{graph}\}\),
\[
z_{g,i,k}=\frac{r_{g,i,k}-\mu_{g,k}}{\sigma_{g,k}+\epsilon},
\]
where \(\mu_{g,k}\) and \(\sigma_{g,k}\) are the mean and standard deviation of reward \(k\) across the \(K\) candidates.

A reward that is constant across the group has zero-centered values and therefore contributes zero to every candidate. We combine the standardized rewards as
\[
s_{g,i}=\sum_k\lambda_k z_{g,i,k}, \qquad \boldsymbol{\lambda}=(0.60,0.60,0.15,0.15)^\top,
\]
corresponding to the fixed relative ratio 4:4:1:1. The same ratio is used for every backbone and track. The aggregate is then standardized over the optimization batch:
\[
A_{g,i}=\frac{s_{g,i}-\mu_{\mathcal{B}}}{\sigma_{\mathcal{B}}+\epsilon},
\]
where \(\mu_{\mathcal{B}}\) and \(\sigma_{\mathcal{B}}\) are computed over all candidate scores in the batch. The resulting caption-level advantage \(A_{g,i}\) is shared by all response tokens in candidate \(i\). Token-level policy ratios, clipping, and KL regularization follow the original GDPO objective without modification; only the reward vector and aggregation are changed.

\section{Experiments}

\begin{table*}[t]
\centering

{\small
\setlength{\tabcolsep}{1mm}
\renewcommand{\arraystretch}{1.08}

\begin{tabular}{@{}lllcccccccccc@{}}
\toprule
\multirow{2}{*}{Track}
& \multirow{2}{*}{Backbone}
& \multirow{2}{*}{Stage}
& \multicolumn{3}{c}{Aggregate Scores}
& \multicolumn{4}{c}{Relevance Metrics}
& \multicolumn{2}{c}{Factuality Metrics}
& \multirow{2}{*}{\(\Delta\)} \\
\cmidrule(lr){4-6}
\cmidrule(lr){7-10}
\cmidrule(lr){11-12}

&&&
Ovr. & Rel. & Fact.
& BERT & R-1 & Sim. & BLT
& MedCAT & Align
& \\

\midrule

\multirow{6}{*}{Std.}
& \multirow{2}{*}{MedGemma-4B}
& SFT
& 0.3669 & 0.5494 & 0.1843
& 0.6081 & 0.2900 & 0.9671 & 0.3325
& 0.2067 & 0.1619
& -- \\

&
& GDPO
& \textbf{0.3798} & \textbf{0.5664} & \textbf{0.1932}
& \textbf{0.6205} & \textbf{0.3223}
& \textbf{0.9827} & \textbf{0.3402}
& \textbf{0.2189} & \textbf{0.1676}
& +3.54\% \\

\addlinespace[1pt]

&
\multirow{2}{*}{Qwen3.5-MedVL}
& SFT
& 0.3675 & 0.5537 & 0.1812
& 0.6160 & 0.2939 & 0.9715 & 0.3335
& 0.2054 & 0.1570
& -- \\

&
& GDPO
& \textbf{0.3898} & \textbf{0.5790} & \textbf{0.2007}
& \textbf{0.6266} & \textbf{0.3318}
& \textbf{1.0052} & \textbf{0.3525}
& \textbf{0.2314} & \textbf{0.1700}
& +6.09\% \\

\addlinespace[1pt]

&
\multirow{2}{*}{Qwen3-VL-4B}
& SFT
& 0.3685 & 0.5475 & 0.1895
& 0.6104 & 0.2901 & 0.9587 & 0.3307
& 0.2083 & 0.1707
& -- \\

&
& GDPO
& \textbf{0.3809} & \textbf{0.5669} & \textbf{0.1948}
& \textbf{0.6252} & \textbf{0.3222}
& \textbf{0.9784} & \textbf{0.3418}
& \textbf{0.2180} & \textbf{0.1715}
& +3.35\% \\

\midrule

\multirow{6}{*}{Syn.}
& \multirow{2}{*}{MedGemma-4B}
& SFT
& 0.5688 & 0.7042 & 0.4334
& 0.7457 & 0.6174
& \textbf{0.9553} & \textbf{0.4985}
& 0.5124 & 0.3544
& -- \\

&
& GDPO
& \textbf{0.5870} & \textbf{0.7070} & \textbf{0.4670}
& \textbf{0.7531} & \textbf{0.6268}
& 0.9506 & 0.4975
& \textbf{0.5276} & \textbf{0.4063}
& +3.20\% \\

\addlinespace[1pt]

&
\multirow{2}{*}{Qwen3.5-MedVL}
& SFT
& 0.5766 & 0.7112 & 0.4420
& 0.7509 & 0.6254
& \textbf{0.9638} & 0.5047
& 0.5236 & 0.3603
& -- \\

&
& GDPO
& \textbf{0.5870} & \textbf{0.7153} & \textbf{0.4588}
& \textbf{0.7570} & \textbf{0.6351}
& 0.9636 & \textbf{0.5054}
& \textbf{0.5338} & \textbf{0.3838}
& +1.82\% \\

\addlinespace[1pt]

&
\multirow{2}{*}{Qwen3-VL-4B}
& SFT
& 0.5722 & 0.7075 & 0.4370
& 0.7483 & 0.6219
& 0.9564 & \textbf{0.5032}
& 0.5173 & 0.3566
& -- \\

&
& GDPO
& \textbf{0.5846} & \textbf{0.7107} & \textbf{0.4585}
& \textbf{0.7538} & \textbf{0.6298}
& \textbf{0.9571} & 0.5024
& \textbf{0.5249} & \textbf{0.3920}
& +2.17\% \\

\bottomrule
\end{tabular}
}

\caption{Organizer-evaluated hidden test performance.
Bold indicates the better result within each SFT--GDPO pair.
R-1, Sim., and BLT denote ROUGE-1, scaled MedImageInsight
image--caption similarity, and BLEURT, respectively.
\(\Delta\) denotes the relative improvement in Overall score over SFT.}
\label{tab:hidden_test}

\end{table*}

\subsection{Datasets and Evaluation Metrics}

We evaluate the Caption Prediction task on the Standard and Synthetical tracks of ImageCLEFmedical Caption 2026, training and selecting models separately for each track ~\cite{ruckert2024rocov2,damm2026imageclef}. The Standard track contains 97,364 training, 19,240 validation, and 15,249 hidden-test examples, while the Synthetical track contains 97,222, 19,239, and 15,262 examples, respectively. Validation sets are used for checkpoint selection. The Standard validation set is additionally used for reward ablations and post-hoc diagnostics, whereas hidden-test predictions are evaluated by the task organizers.

The official evaluator reports Relevance, Factuality, and Overall. Relevance averages IDF-weighted BERTScore Recall, ROUGE-1 F1, BLEURT, and MedImageInsight image–-caption similarity. Factuality averages MedCAT-based UMLS Concept F1 and AlignScore, with the reference caption as context and the generated caption as the claim. Overall is the mean of Relevance and Factuality.

Some training rewards are related to official evaluation components but use different formulations. BioBERTScore-F1 differs from the official IDF-weighted BERTScore Recall, and ROUGE-SU4 differs from ROUGE-1 F1. Both the neighborhood reward and one Relevance component use MedImageInsight, but the official metric directly scores the target image–caption pair, whereas our reward aligns the reference- and candidate-induced distributions over a common training-image bank. We therefore use the official metrics for benchmark evaluation and the paired structured diagnostics only to analyze the properties targeted by the proposed rewards.

\subsection{Models, Training, and Decoding}

We instantiate the framework with Qwen3.5-2B-MedVL~\cite{openmed_qwen35_2b_medvl}, MedGemma-4B~\cite{medgemma2025}, and Qwen3-VL-4B~\cite{Qwen3-VL}. All model parameters are updated during SFT and GDPO, while the reward-side encoders and graph extractor remain frozen. The same stage-specific hyperparameters and reward weights are used for both tracks. All models use the same medical-captioning instruction; the full prompts and configuration are provided in the supplement.

\subsection{Baselines and Evaluation Protocol}

For each backbone and track, the selected SFT checkpoint initializes GDPO and serves as the matched baseline. SFT and GDPO use the same architecture, prompt, validation-based selection criterion, and hidden-test decoding configuration. The six matched comparisons therefore evaluate the complete post-SFT procedure relative to its own SFT initialization across model families and task settings. They do not compare GDPO with alternative policy-optimization algorithms; GDPO is held fixed, and the study focuses on the proposed reward objective.

Reward ablations are conducted with Qwen3.5-2B-MedVL on the Standard validation set. Single-reward variants retain one component, while leave-one-out variants remove one component and preserve the original relative weights of the others. All variants use the same checkpoint-saving and validation-selection protocol. Because each ablation configuration is based on one training run and several differences are small, the results are interpreted descriptively rather than as statistically resolved rankings among reward combinations.

Post-hoc diagnostics compare the selected SFT and GDPO checkpoints on all 19,240 Standard validation examples using greedy decoding. The diagnostics are performed after model selection and do not affect checkpoint choice. We report image-neighborhood KL divergence, entity consistency, aggregate graph consistency, and relation consistency, with 95\% confidence intervals obtained from 10,000 paired bootstrap resamples. These analyses use the same frozen reward-side representations and are intended to examine whether optimization changes the structured properties targeted by the objective, not to provide independent clinical verification.

\begin{table}[t]
\centering
\small
\setlength{\tabcolsep}{3.2pt}
\begin{tabular}{lccccc}
\toprule
Configuration & Step & Rel. & Fact. & Ovr. & \(\Delta\) \\
\midrule
Full          & 1000 & 0.5722 & 0.1834 & \textbf{0.3778} & -- \\
\midrule
Semantic only &  750 & 0.5664 & \textbf{0.1861} & 0.3763 & -0.0016 \\
Lexical only  &  250 & 0.5557 & 0.1751 & 0.3654 & -0.0124 \\
Neighborhood only &  250 & 0.5475 & 0.1500 & 0.3487 & -0.0291 \\
Graph only    & 1250 & 0.5403 & 0.1751 & 0.3577 & -0.0201 \\
\midrule
w/o Semantic  & 1750 & 0.5721 & 0.1819 & 0.3770 & -0.0009 \\
w/o Lexical   & 1000 & \textbf{0.5746} & 0.1726 & 0.3736 & -0.0042 \\
w/o Neighborhood  &  750 & 0.5708 & 0.1789 & 0.3749 & -0.0030 \\
w/o Graph     & 1750 & 0.5737 & 0.1732 & 0.3734 & -0.0044 \\
\bottomrule
\end{tabular}
\caption{
Reward ablation on the Standard validation set using Qwen3.5-MedVL. Neighborhood denotes the distributional image-neighborhood
reward \(r_{\mathrm{nbr}}\). \(\Delta\) denotes the signed Overall difference relative to the full objective and is computed from unrounded scores.
}
\label{tab:reward_ablation}
\end{table}

\section{Results}
\begin{figure}[t]
    \centering
    \includegraphics[width=\columnwidth]
    {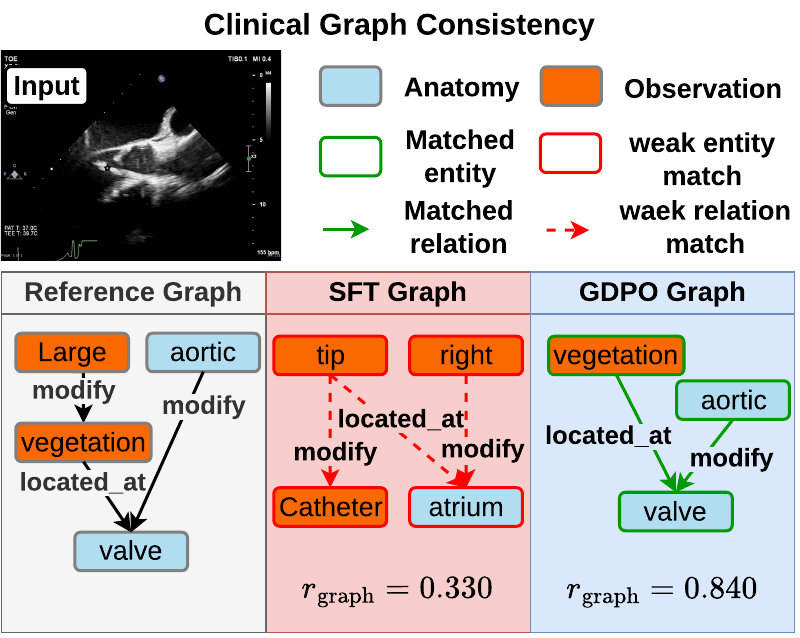}
    \caption{
         Clinical graph consistency, where GDPO corrects the finding and
        recovers its localization relation.
        The supplementary material provides 
        additional successful, failed, and ambiguous cases.;CC-BY [Muacevic et al(2024)]
    }
    \label{fig:reward_diagnostics}
\end{figure}

% \begin{figure*}[t]
%     \centering
%     \includegraphics[width=\textwidth]
%     {Figures/structured_reward_diagnostics.pdf}
%     \caption{
%         Selected successful cases illustrating the two structured rewards.
%         (A) image-neighborhood alignment, where the GDPO caption shifts the
%         caption-induced retrieval distribution toward the reference-induced
%         image neighborhood.
%         (B) Clinical graph consistency, where GDPO corrects the finding and
%         recovers its localization relation.
%         The supplementary material provides 
%         additional successful, failed, and ambiguous cases.
%         \textit{Image attribution (all CC BY):}
%         In (A), the input image is attributed to Muacevic et al. (2024);
%         Reference/GDPO retrievals 1--3 are attributed to Muacevic et al. (2023),
%         Baig et al. (2023), and Muacevic et al. (2021), respectively;
%         SFT retrievals 1--3 are attributed to da Silva et al. (2017),
%         Wang et al. (2013), and Aurilio et al. (2010), respectively.
%         The image in (B) is attributed to Muacevic et al. (2024).}
%     \label{fig:reward_diagnostics}
% \end{figure*}

\subsection{Hidden test Performance}

Table~\ref{tab:hidden_test} reports organizer-evaluated performance on the hidden test sets. The proposed GDPO stage improves Overall, Relevance, and Factuality over the matched SFT baseline in all six backbone–track combinations. Averaged across the three backbones and two tracks, the relative improvements are 3.36\% in Overall, 2.11\% in Relevance, and 5.81\% in Factuality. These results indicate that the complete post-SFT procedure consistently improves its own SFT initialization across different model families and data regimes.

On the Standard track, Qwen3.5-2B-MedVL shows the largest gain, improving Overall from 0.3675 to 0.3898, corresponding to a relative improvement of 6.09\%. Its Relevance and Factuality scores increase from 0.5537 to 0.5790 and from 0.1812 to 0.2007, respectively. MedGemma-4B improves from 0.3669 to 0.3798, while Qwen3-VL-4B improves from 0.3685 to 0.3809. Thus, the benefit is not confined to a single backbone. On the Synthetical track, Overall improves from 0.5688 to 0.5870 for MedGemma-4B, from 0.5766 to 0.5870 for Qwen3.5-2B-MedVL, and from 0.5722 to 0.5846 for Qwen3-VL-4B. The Synthetical-track gains are driven primarily by Factuality, whereas Relevance changes more modestly and some individual component metrics remain nearly unchanged or decrease slightly. The aggregate improvements therefore do not arise from uniform increases in every official metric.

These comparisons should be interpreted as matched evaluations of the complete SFT-to-GDPO procedure. Architecture, prompts, validation-based checkpoint selection, and hidden-test decoding are held fixed within each pair. The results show that post-SFT optimization with the proposed objective improves the corresponding SFT model, but they do not isolate the reward design from the general effect of applying GDPO instead of stopping after SFT.

\subsection{Reward Component Ablation}

Table~\ref{tab:reward_ablation} analyzes the four reward components on the Standard validation set using Qwen3.5-2B-MedVL. The full objective achieves the highest Overall score of 0.3778. Among the single-reward variants, semantic fidelity is the strongest, reaching 0.3763 Overall and 0.1861 Factuality. Lexical fidelity alone reaches 0.3654 Overall, while the neighborhood-only and graph-only variants reach 0.3487 and 0.3577, respectively. These results show that the structured rewards are weak as standalone objectives under the tested setting and require the stronger reference-based signals to maintain overall caption quality.

The leave-one-out results provide a complementary view. Removing semantic fidelity reduces Overall from 0.3778 to 0.3770, while removing lexical fidelity lowers it to 0.3736. Removing the neighborhood reward reduces Overall to 0.3749 and Factuality from 0.1834 to 0.1789. Removing the graph reward yields 0.3734 Overall and 0.1732 Factuality. Although the neighborhood and graph rewards perform poorly in isolation, removing either from the full objective degrades the selected model, particularly on Factuality. This pattern supports their intended role as complementary structural signals rather than replacements for semantic and lexical fidelity.

All ablation variants follow the same checkpoint-saving and validation-based selection protocol, but each configuration is based on a single training run and several differences are small. The results should therefore be interpreted descriptively. They support the overall complementarity of the reward components but do not establish statistically resolved rankings among individual reward combinations.

\subsection{Post-hoc Structured Reward Diagnostics}

\begin{table}[t]
\centering

\begingroup
\small
\setlength{\tabcolsep}{1.5pt}
\begin{tabular}{@{}lccc@{}}
\toprule
Metric
& SFT \(\rightarrow\) GDPO
& \(\Delta\)
& 95\% CI \\
\midrule
Nbr. KL \(\downarrow\)
& \(0.2323 \rightarrow 0.2176\)
& \(-0.0147\)
& \([-0.0158,-0.0135]\) \\
Entity \(F_E\) \(\uparrow\)
& \(0.3629 \rightarrow 0.3904\)
& \(+0.0275\)
& \([0.0247,0.0303]\) \\
Graph reward \(\uparrow\)
& \(0.3010 \rightarrow 0.3244\)
& \(+0.0234\)
& \([0.0208,0.0259]\) \\
Relation \(F_R\) \(\uparrow\)
& \(0.1688 \rightarrow 0.1838\)
& \(+0.0150\)
& \([0.0123,0.0177]\) \\
\bottomrule
\end{tabular}
\endgroup

\caption{
Paired post-hoc diagnostics comparing the selected SFT and GDPO
checkpoints under greedy decoding on the Standard validation set.
Nbr. KL denotes
\(D_{\mathrm{KL}}(p^*\Vert q)\).
Entries report mean per-example scores, with
\(\Delta=\mathrm{GDPO}-\mathrm{SFT}\); 95\% confidence intervals
are obtained from 10,000 paired bootstrap resamples of the
per-example differences.
Entity and graph metrics are computed on references with at least
one extracted entity (16,699/19,240; 86.8\%), and relation
consistency on references with at least one extracted relation
(15,196/19,240; 79.0\%).
}
\label{tab:reward_diagnostics}
\end{table}

We next compare the selected SFT and GDPO checkpoints on all 19,240 Standard validation examples using greedy decoding. These diagnostics are conducted after checkpoint selection and do not affect model choice. They use the same frozen MedImageInsight, RadGraph-XL, and SapBERT representations employed by the reward functions, so they are mechanism-oriented analyses rather than independent clinical evaluations.

Table~\ref{tab:reward_diagnostics} shows favorable mean changes in all four diagnostics. Image-neighborhood divergence decreases from 0.2323 to 0.2176, corresponding to a mean change of -0.0147 with a paired 95\% bootstrap confidence interval of [-0.0158, -0.0135]. Entity consistency increases from 0.3629 to 0.3904, graph consistency from 0.3010 to 0.3244, and relation consistency from 0.1688 to 0.1838. Their mean changes are +0.0275, +0.0234, and +0.015, respectively, and all example-level confidence intervals exclude zero. The graph metrics apply to 16,699 of the 19,240 references, or 86.8\%, containing at least one extracted entity, while relation consistency applies to 15,196 references, or 79.0\%, containing at least one extracted relation.

The improvements are not uniform across examples. Neighborhood KL decreases for 57.01\% of cases and increases for 42.80\%, while graph consistency improves for 53.79\% and regresses for 42.33\%. Thus, the proposed objective shifts average structured behavior in the intended direction without improving every example. Per-example neighborhood-KL reductions correlate with gains in the official image-similarity score at Spearman’s \(\rho\)=0.289, whereas graph-consistency gains correlate more weakly with official Factuality gains at \(\rho\)=0.132. The first association is partly expected because both measures use MedImageInsight representations. The weaker graph correlation is also reasonable because the graph diagnostic captures reference-conditioned entity, assertion, and relation agreement, whereas the official Factuality score combines concept overlap and textual entailment.

The reported confidence intervals characterize variation across validation examples for the selected checkpoints, not variation across independent training runs. Accordingly, these analyses support the intended mechanism of the structured rewards but do not demonstrate training-level statistical superiority or complete image-grounded clinical correctness.

\subsection{Qualitative Analysis}
Figure~\ref{fig:reward_diagnostics} further illustrates the effect of graph consistency reward. GDPO corrects the predicted finding from a catheter-tip description to an aortic-valve vegetation and recovers the corresponding localization relation. This example is illustrative rather than representative; additional successful, failed, and ambiguous cases are provided in the supplementary material.

\section{Conclusion}

We introduced two structured surrogate rewards for post-SFT medical image captioning: distributional image-neighborhood alignment and semantic one-to-one matching of clinical entities, assertions, and relations. Across three backbones and two ImageCLEFmedical Caption tracks, GDPO consistently improved matched SFT baselines. Ablations indicate that the structured rewards complement stronger semantic and lexical signals, while paired diagnostics show favorable changes in the targeted neighborhood and graph representations. These rewards remain reference-conditioned and inherit biases and extraction errors from their frozen reward models; they do not independently verify image-grounded clinical correctness or validate the system for clinical use. Future work should evaluate broader clinical-report settings, reward-model reliability, training-run variability, and adaptive reward weighting.

\section*{Acknowledgments}

We thank the ImageCLEFmedical Caption 2026 organizers for
providing the benchmark, evaluation framework, and official
evaluation results. This work was supported in part by the
National Research Foundation of Korea (NRF) grant funded by
the Korean government (MSIT) (grant number:
RS-2024-00360176), and in part by the Korea Health Technology
R\&D Project through the Korea Health Industry Development
Institute (KHIDI), funded by the Ministry of Health and Welfare,
Republic of Korea (grant number: RS-2025-02222326).

\bibliography{aaai2027}

@article{jain2021radgraph,
  title={Radgraph: Extracting clinical entities and relations from radiology reports},
  author={Jain, Saahil and Agrawal, Ashwin and Saporta, Adriel and Truong, Steven QH and Duong, Du Nguyen and Bui, Tan and Chambon, Pierre and Zhang, Yuhao and Lungren, Matthew P and Ng, Andrew Y and others},
  journal={arXiv preprint arXiv:2106.14463},
  year={2021}
}

@inproceedings{delbrouck2024radgraph,
  title={Radgraph-xl: A large-scale expert-annotated dataset for entity and relation extraction from radiology reports},
  author={Delbrouck, Jean-Benoit and Chambon, Pierre and Chen, Zhihong and Varma, Maya and Johnston, Andrew and Blankemeier, Louis and Van Veen, Dave and Bui, Tan and Truong, Steven and Langlotz, Curtis},
  booktitle={Findings of the Association for Computational Linguistics: ACL 2024},
  pages={12902--12915},
  year={2024}
}

@inproceedings{rennie2017selfcritical,
  title     = {Self-Critical Sequence Training for Image Captioning},
  author    = {Rennie, Steven J. and Marcheret, Etienne and
               Mroueh, Youssef and Ross, Jarret and Goel, Vaibhava},
  booktitle = {Proceedings of the IEEE Conference on Computer Vision
               and Pattern Recognition},
  year      = {2017}
}

@inproceedings{lin2004rouge,
  title     = {{ROUGE}: A Package for Automatic Evaluation of Summaries},
  author    = {Lin, Chin-Yew},
  booktitle = {Text Summarization Branches Out},
  pages     = {74--81},
  address   = {Barcelona, Spain},
  publisher = {Association for Computational Linguistics},
  year      = {2004},
  url       = {https://aclanthology.org/W04-1013/}
}

@inproceedings{zhang2020bertscore,
  title     = {{BERTScore}: Evaluating Text Generation with {BERT}},
  author    = {Zhang, Tianyi and Kishore, Varsha and Wu, Felix and
               Weinberger, Kilian Q. and Artzi, Yoav},
  booktitle = {International Conference on Learning Representations},
  year      = {2020}
}

@article{lee2020biobert,
  title     = {{BioBERT}: A Pre-trained Biomedical Language
               Representation Model for Biomedical Text Mining},
  author    = {Lee, Jinhyuk and Yoon, Wonjin and Kim, Sungdong and
               Kim, Donghyeon and Kim, Sunkyu and So, Chan Ho and
               Kang, Jaewoo},
  journal   = {Bioinformatics},
  volume    = {36},
  number    = {4},
  pages     = {1234--1240},
  year      = {2020},
  doi       = {10.1093/bioinformatics/btz682}
}

@article{codella2024medimageinsight,
  title     = {{MedImageInsight}: An Open-Source Embedding Model for
               General Domain Medical Imaging},
  author    = {Codella, Noel C. F. and Jin, Ying and Jain, Shrey and
               Gu, Yu and Lee, Ho Hin and Ben Abacha, Asma and
               Santamaria-Pang, Alberto and Guyman, Will and
               Sangani, Naiteek and Zhang, Sheng and Poon, Hoifung and
               Hyland, Stephanie and Bannur, Shruthi and
               Alvarez-Valle, Javier and Li, Xue and Garrett, John and
               McMillan, Alan and Rajguru, Gaurav and Maddi, Madhu and
               Vijayrania, Nilesh and Bhimai, Rehaan and
               Mecklenburg, Nick and Jain, Rupal and Holstein, Daniel and
               Gaur, Naveen and Aski, Vijay and Hwang, Jenq-Neng and
               Lin, Thomas and Tarapov, Ivan and Lungren, Matthew and
               Wei, Mu},
  journal   = {arXiv preprint arXiv:2410.06542},
  year      = {2024},
  doi       = {10.48550/arXiv.2410.06542}
}

@inproceedings{liu2021sapbert,
  title     = {Self-Alignment Pretraining for Biomedical Entity
               Representations},
  author    = {Liu, Fangyu and Shareghi, Ehsan and Meng, Zaiqiao and
               Basaldella, Marco and Collier, Nigel},
  booktitle = {Proceedings of the 2021 Conference of the North American
               Chapter of the Association for Computational Linguistics:
               Human Language Technologies},
  pages     = {4228--4238},
  publisher = {Association for Computational Linguistics},
  year      = {2021},
  doi       = {10.18653/v1/2021.naacl-main.334}
}

@article{liu2026gdpo,
  title     = {{GDPO}: Group Reward-Decoupled Normalization Policy
               Optimization for Multi-Reward {RL} Optimization},
  author    = {Liu, Shih-Yang and Dong, Xin and Lu, Ximing and
               Diao, Shizhe and Belcak, Peter and Liu, Mingjie and
               Chen, Min-Hung and Yin, Hongxu and
               Wang, Yu-Chiang Frank and Cheng, Kwang-Ting and
               Choi, Yejin and Kautz, Jan and Molchanov, Pavlo},
  journal   = {arXiv preprint arXiv:2601.05242},
  year      = {2026}
}

@inproceedings{damm2026imageclef,

  author = {Damm, Hendrik and Pakull, Tabea M. G. and Reinartz, Lea and Bracke, Benjamin and Ery{\i}lmaz, Bahad{\i}r and Nath, Praveen and Br{\"u}ngel, Raphael and Schmidt, Cynthia S. and Sch{\"a}fer, Henning and Ben Abacha, Asma and {Garc{\'\i}a Seco de Herrera}, Alba and M{\"u}ller, Henning and Friedrich, Christoph M.},

  title = {Overview of {ImageCLEFmedical} 2026 -- Medical Concept Detection and Caption Generation with Synthetic Data Extensions},

  booktitle = {CLEF2026 Working Notes},

  series = {{CEUR} Workshop Proceedings},

  year = {2026},

  publisher = {CEUR-WS.org},

  month = {September 21-24},

  address = {Jena, Germany},

  note = {To appear}

}

@article{ruckert2024rocov2,

 title = {{ROCOv2}: Radiology Objects in COntext Version 2, an Updated Multimodal Image Dataset},

 author = {Johannes R{\"u}ckert and Louise Bloch and Raphael Br{\"u}ngel and Ahmad Idrissi{-}Yaghir and Henning Sch{\"a}fer and Cynthia S. Schmidt and Sven Koitka and Obioma Pelka and Asma Ben Abacha and Alba Garc{'{\i}}a Seco de Herrera and Henning M{\"u}ller and Peter Horn and Felix Nensa and Christoph M. Friedrich},

 journal = {Scientific Data},

 volume = {11},

 number = {1},

 year = {2024},

 doi = {10.1038/s41597-024-03496-6}

}

@article{johnson2019mimic,
  title     = {{MIMIC-CXR}, a De-identified Publicly Available
               Database of Chest Radiographs with Free-text Reports},
  author    = {Johnson, Alistair E. W. and Pollard, Tom J. and
               Berkowitz, Seth J. and Greenbaum, Nathaniel R. and
               Lungren, Matthew P. and Deng, Chih-ying and
               Mark, Roger G. and Horng, Steven},
  journal   = {Scientific Data},
  volume    = {6},
  number    = {1},
  pages     = {317},
  year      = {2019},
  doi       = {10.1038/s41597-019-0322-0}
}

@article{demnerfushman2016iu,
  title     = {Preparing a Collection of Radiology Examinations
               for Distribution and Retrieval},
  author    = {Demner-Fushman, Dina and Kohli, Marc D. and
               Rosenman, Marc B. and Shooshan, Sonya E. and
               Rodriguez, Laritza and Antani, Sameer and
               Thoma, George R. and McDonald, Clement J.},
  journal   = {Journal of the American Medical Informatics Association},
  volume    = {23},
  number    = {2},
  pages     = {304--310},
  year      = {2016},
  doi       = {10.1093/jamia/ocv080}
}

@article{xiao2025mpo,
  title     = {Radiology Report Generation via Multi-objective
               Preference Optimization},
  author    = {Xiao, Ting and Shi, Lei and Liu, Peng and
               Wang, Zhe and Bai, Chenjia},
  journal   = {Proceedings of the AAAI Conference on Artificial Intelligence},
  volume    = {39},
  number    = {8},
  pages     = {8664--8672},
  year      = {2025},
  doi       = {10.1609/aaai.v39i8.32936}
}

@article{wang2026himed,
  title     = {Beyond N-grams: A Hierarchical Reward Learning Framework
               for Clinically-Aware Medical Report Generation},
  author    = {Wang, Yuan and Gao, Shujian and Liu, Jiaxiang and
               Jiang, Songtao and Xia, Haoxiang and Zhang, Xiaotian and
               Kang, Zhaolu and Wang, Yemin and Liu, Zuozhu},
  journal   = {Proceedings of the AAAI Conference on Artificial Intelligence},
  volume    = {40},
  number    = {40},
  pages     = {33719--33727},
  year      = {2026},
  doi       = {10.1609/aaai.v40i40.40662}
}

@InProceedings{chen2026orapo,
    author    = {Chen, Zhuoxiao and Yu, Hongyang and Xu, Ying and Luo, Yadan and Duong, Long and Li, Yuan-Fang},
    title     = {OraPO: Oracle-educated Reinforcement Learning for Data-efficient and Factual Radiology Report Generation},
    booktitle = {Proceedings of the IEEE/CVF Conference on Computer Vision and Pattern Recognition (CVPR)},
    month     = {June},
    year      = {2026},
    pages     = {28275-28287}
}

@inproceedings{delbrouck2022semantic,
    title = "Improving the Factual Correctness of Radiology Report Generation with Semantic Rewards",
    author = "Delbrouck, Jean-Benoit  and
      Chambon, Pierre  and
      Bluethgen, Christian  and
      Tsai, Emily  and
      Almusa, Omar  and
      Langlotz, Curtis",
    editor = "Goldberg, Yoav  and
      Kozareva, Zornitsa  and
      Zhang, Yue",
    booktitle = "Findings of the Association for Computational Linguistics: EMNLP 2022",
    month = dec,
    year = "2022",
    address = "Abu Dhabi, United Arab Emirates",
    publisher = "Association for Computational Linguistics",
    url = "https://aclanthology.org/2022.findings-emnlp.319/",
    doi = "10.18653/v1/2022.findings-emnlp.319",
    pages = "4348--4360"
}

@inproceedings{sellam2020bleurt,
  title     = {{BLEURT}: Learning Robust Metrics for Text Generation},
  author    = {Sellam, Thibault and Das, Dipanjan and Parikh, Ankur P.},
  booktitle = {Proceedings of the 58th Annual Meeting of the
               Association for Computational Linguistics},
  pages     = {7881--7892},
  year      = {2020},
  doi       = {10.18653/v1/2020.acl-main.704}
}

@inproceedings{zha2023alignscore,
  title     = {{AlignScore}: Evaluating Factual Consistency with
               a Unified Alignment Function},
  author    = {Zha, Yuheng and Yang, Yichi and Li, Ruichen and Hu, Zhiting},
  booktitle = {Proceedings of the 61st Annual Meeting of the
               Association for Computational Linguistics},
  pages     = {11328--11348},
  year      = {2023},
  doi       = {10.18653/v1/2023.acl-long.634}
}

@inproceedings{kim2026gdpo,
  author    = {Kim, Hyun Jun and Shin, Heeseung and Kim, Minjun and Lim, Changwon},
  title     = {{GDPO}-Based Multi-Alignment Reward Optimization for
               Clinically Grounded Medical Image Captioning},
  booktitle = {CLEF 2026 Working Notes},
  year      = {2026},
  note      = {Accepted; to appear in CEUR Workshop Proceedings}
}

@inproceedings{bengio2015scheduled,
  title     = {Scheduled Sampling for Sequence Prediction with
               Recurrent Neural Networks},
  author    = {Bengio, Samy and Vinyals, Oriol and Jaitly, Navdeep
               and Shazeer, Noam},
  booktitle = {Advances in Neural Information Processing Systems},
  volume    = {28},
  year      = {2015}
}

@inproceedings{ranzato2016sequence,
  title     = {Sequence Level Training with Recurrent Neural Networks},
  author    = {Ranzato, Marc'Aurelio and Chopra, Sumit and
               Auli, Michael and Zaremba, Wojciech},
  booktitle = {International Conference on Learning Representations},
  year      = {2016}
}

@ARTICLE{kraljevic2019medcat,
  title="Multi-domain clinical natural language processing with {MedCAT}: The Medical Concept Annotation Toolkit",
  author="Kraljevic, Zeljko and Searle, Thomas and Shek, Anthony and Roguski, Lukasz and Noor, Kawsar and Bean, Daniel and Mascio, Aurelie and Zhu, Leilei and Folarin, Amos A and Roberts, Angus and Bendayan, Rebecca and Richardson, Mark P and Stewart, Robert and Shah, Anoop D and Wong, Wai Keong and Ibrahim, Zina and Teo, James T and Dobson, Richard J B",
  journal="Artif. Intell. Med.",
  volume=117,
  pages="102083",
  month=jul,
  year=2021,
  issn="0933-3657",
  doi="10.1016/j.artmed.2021.102083"
}

@inproceedings{cho2023finegrainedimagecaptioningclip,
    title = "Fine-grained Image Captioning with {CLIP} Reward",
    author = "Cho, Jaemin  and
      Yoon, Seunghyun  and
      Kale, Ajinkya  and
      Dernoncourt, Franck  and
      Bui, Trung  and
      Bansal, Mohit",
    editor = "Carpuat, Marine  and
      de Marneffe, Marie-Catherine  and
      Meza Ruiz, Ivan Vladimir",
    booktitle = "Findings of the Association for Computational Linguistics: NAACL 2022",
    month = jul,
    year = "2022",
    address = "Seattle, United States",
    publisher = "Association for Computational Linguistics",
    url = "https://aclanthology.org/2022.findings-naacl.39/",
    doi = "10.18653/v1/2022.findings-naacl.39",
    pages = "517--527"
}

@article{Qwen3-VL,
      title={Qwen3-VL Technical Report}, 
      author={Shuai Bai and Yuxuan Cai and Ruizhe Chen and Keqin Chen and Xionghui Chen and Zesen Cheng and Lianghao Deng and Wei Ding and Chang Gao and Chunjiang Ge and Wenbin Ge and Zhifang Guo and Qidong Huang and Jie Huang and Fei Huang and Binyuan Hui and Shutong Jiang and Zhaohai Li and Mingsheng Li and Mei Li and Kaixin Li and Zicheng Lin and Junyang Lin and Xuejing Liu and Jiawei Liu and Chenglong Liu and Yang Liu and Dayiheng Liu and Shixuan Liu and Dunjie Lu and Ruilin Luo and Chenxu Lv and Rui Men and Lingchen Meng and Xuancheng Ren and Xingzhang Ren and Sibo Song and Yuchong Sun and Jun Tang and Jianhong Tu and Jianqiang Wan and Peng Wang and Pengfei Wang and Qiuyue Wang and Yuxuan Wang and Tianbao Xie and Yiheng Xu and Haiyang Xu and Jin Xu and Zhibo Yang and Mingkun Yang and Jianxin Yang and An Yang and Bowen Yu and Fei Zhang and Hang Zhang and Xi Zhang and Bo Zheng and Humen Zhong and Jingren Zhou and Fan Zhou and Jing Zhou and Yuanzhi Zhu and Ke Zhu},
	  journal={arXiv preprint arXiv:2511.21631},
      year={2025}
}

@misc{openmed_qwen35_2b_medvl,
  title        = {{OpenMed/Qwen3.5-2B-MedVL}},
  author       = {{OpenMed}},
  year         = {2026},
  howpublished = {\url{https://huggingface.co/OpenMed/Qwen3.5-2B-MedVL}},
  note         = {Hugging Face model card, accessed 2026-05-19}
}

@article{medgemma2025,
  title={MedGemma Technical Report},
  author={Sellergren, Andrew and Kazemzadeh, Sahar and Jaroensri, Tiam and Kiraly, Atilla and Traverse, Madeleine and Kohlberger, Timo and Xu, Shawn and Jamil, Fayaz and Hughes, Cían and Lau, Charles and others},
  journal={arXiv preprint arXiv:2507.05201},
  year={2025}
}

@article{li2026claimdiff,
  title={ClaimDiff-RL: Fine-Grained Caption Reinforcement Learning through Visual Claim Comparison},
  author={Li, Tianle and Shen, Xuyang and Ma, Yan and Guo, Rongxin and Chen, Shaoxiang and Chen, Jiacheng and Wang, Haochen and Tang, Hongyang and Zhou, Yucong and Cheng, Yu},
  journal={arXiv preprint arXiv:2605.20278},
  year={2026}
}

@inproceedings{zhang2025sc,
  title={Sc-captioner: Improving image captioning with self-correction by reinforcement learning},
  author={Zhang, Lin and Zeng, Xianfang and Li, Kangcong and Yu, Gang and Chen, Tao},
  booktitle={Proceedings of the IEEE/CVF International Conference on Computer Vision},
  pages={23145--23155},
  year={2025}
}

@article{tang2026cccaption,
  title={CCCaption: Dual-Reward Reinforcement Learning for Complete and Correct Image Captioning},
  author={Tang, Zhijiang and Wang, Linhua and Qi, Jiaxin and Jiang, Weihao and Hou, Peng and Zeng, Anxiang and Huang, Jianqiang},
  journal={arXiv preprint arXiv:2602.21655},
  year={2026}
}


\end{document}